\documentclass[acmtog]{acmart}
\usepackage{listings}
\usepackage{xcolor}
\usepackage{graphicx}
\usepackage{booktabs} % For better looking tables
\usepackage{multirow} % For multirow cells

\AtBeginDocument{%
  }

\begin{document}

\title{Multimodal Language Models for Domain-Specific Procedural Video Summarization}

\author{Nafisa Hussain}
\affiliation{%
  \institution{University of California, Santa Cruz}
  \city{Santa Cruz}
  \country{USA}}
\email{nahussai@ucsc.edu}

\renewcommand{\shortauthors}{Nafisa Hussain}

\begin{abstract}
  Videos serve as a powerful medium to convey ideas, tell stories, and provide detailed instructions, especially through long-format tutorials. Such tutorials are valuable for learning new skills at one's own pace, yet they can be overwhelming due to their length and dense content. Viewers often seek specific information, like precise measurements or step-by-step execution details, making it essential to extract and summarize key segments efficiently. An intelligent, time-sensitive video assistant capable of summarizing and detecting highlights in long videos is highly sought after. Recent advancements in Multimodal Large Language Models offer promising solutions to develop such an assistant. Our research explores the use of multimodal models to enhance video summarization and step-by-step instruction generation within specific domains. These models need to understand temporal events and relationships among actions across video frames. Our approach focuses on fine-tuning TimeChat to improve its performance in specific domains: cooking and medical procedures. By training the model on domain-specific datasets like Tasty for cooking and MedVidQA for medical procedures, we aim to enhance its ability to generate concise, accurate summaries of instructional videos. We curate and restructure these datasets to create high-quality video-centric instruction data. Our findings indicate that when finetuned on domain-specific procedural data, TimeChat can significantly improve the extraction and summarization of key instructional steps in long-format videos. This research demonstrates the potential of specialized multimodal models to assist with practical tasks by providing personalized, step-by-step guidance tailored to the unique aspects of each domain.
\end{abstract}




\maketitle

\section{Introduction}
Various ideas and concepts can be expressed through the medium of videos. Longer format videos provide the means to tell stories, teach topics, and provide detailed instructions. The means to provide in-depth tutorials through visual demonstrations of steps in succession to aid in the visual understanding of procedures greatly benefits viewers whose intent is to learn new skills and tasks at their own time and pace. However, long video tutorials oftentimes can be lengthy and dense in content, making it both time-consuming and frustrating for individuals to sift through for key instructions. The information needs of viewers can vary from looking for correct measurements of materials to understanding tactically how to execute a step. Users' needs can be satisfied by localizing and retrieving video highlights and concisely summarizing the activity or event from highlighted clips. An intelligent, time-sensitive video assistant capable of summarizing and detecting highlights within long-format videos is a longstanding pursuit for developers. Advancements in Large Language Model capabilities show promise in developing such a capable video assistant. 

Many endeavors have been made to incorporate video encoders with LLMs, producing multimodal models capable of basic video understanding. Multimodal models have found application in diverse tasks, illustrating their flexibility and the breadth of their utility. These models have been successfully employed in generating step-by-step instructions for complex tasks. This involves understanding a text-based query and analyzing accompanying visual data to produce a coherent, easily understandable sequence of instructions. A multimodal model must have sufficient temporal understanding of events and actions occurring within video inputs. Models should also be able to capture a detailed understanding of the relationships among different actions presented across a certain window of frames. Additionally, the reasoning ability to work through the relationships between different tools, ingredients, or materials is necessary for meaningful understanding of inputted tasks. This capability represents a significant leap forward in models’ potential to assist with practical, everyday tasks, offering personalized guidance that considers the unique aspects of each scenario.

With the use of these models expanding across domains and into higher complexity tasks, we wish to explore utilizing these models in domain-specific step-by-step procedure understanding. We propose a specialized approach to finetune and enhance the video summarization abilities of Large Vision Language Models for step-by-step instruction generation across different domains. Through the selected domains of cooking and medical procedure summarization, we finetune multimodal language model TimeChat to \textbf{(1)} learn instructional video explanations and event boundaries and \textbf{(2)} summarize the primary procedural activities and their dependencies between clips concisely \cite{Ren2024}. We specifically tune TimeChat on instruction data within the recipe and medical video domain to generate succinct instructional descriptions of health procedures and cooking steps. We additionally explore transferring these learned skills between the recipe video and medical procedural video domains.

\begin{figure}[ht]
    \centering
    \includegraphics[width=8.5cm, height=7cm]{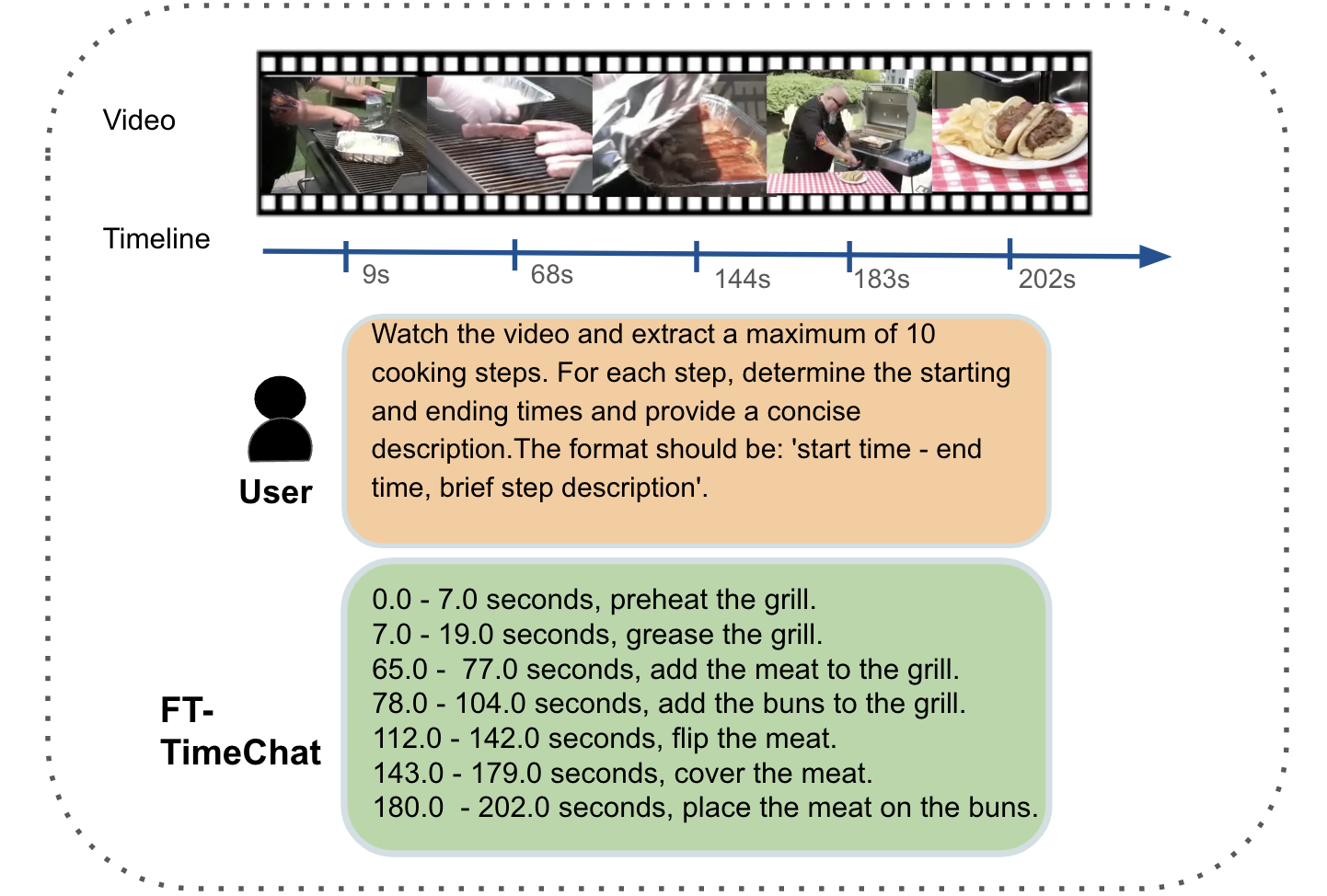}
    \caption{Finetuned TimeChat on Cooking Domain Video Summarization}
    \Description[fig:diagram]{Finetuned TimeChat extracting recipe steps}
    \label{fig:diagram1}
\end{figure}

\section{Related Work}

\subsection{ Multimodal Models (VLMS)}
Advancements in integrating video and time-aware encoders into the architectures of LLMs have given rise to models with multimodal capabilities. Multimodality opens new opportunities for various forms of input and outputs, such as video, audio, and timestamps. Large Vision Language Models (LVLMs) merge computer vision and natural language processing techniques to construct holistic comprehension and reasoning around visual, auditory, and textual information. Many endeavors have been made to integrate video encoders with LLMs for basic video question answering and detailed captioning \cite{Jin2023} \cite{Liu2023} \cite{Song2023}. Many of these methods focus primarily on general video understanding skills like spatiotemporal reasoning and causal relationship inference among simultaneous events. Video-ChatGPT is an LVLM extension of the GPT architecture, leveraging LLaMA as its foundation \cite{Maaz2023}. It employs CLIP ViT-L/14 as its visual encoder and Vicuna as its large language model, the same encoder and decoders used in Video-LLaVA \cite{Radford2021}.  However, when testing Video-ChatGPT's performance in recipe generation, it does not accurately understand the steps and ingredients in the provided cooking videos. The model exhibited poorer performance in summarizing a health-related video that dealt with abstract and more technically prose-heavy concepts and jargon. Despite these shortcomings, these model architectures do show promise in utilizing their language abilities to describe activities and scenarios. The natural language abilities of the LLMs used in these models have the potential to compile and organize information into meaningful and comprehensible instructions.

\subsection{Domains}
Video comprehension through multimodal models has been applied to several domains and formats of videos. Many general comprehension datasets are available to train models for general world knowledge of activities, fictional plot understanding, first-person perspective actions, and so on \cite{Lei2018} \cite{Fan2019} \cite{Li2020} \cite{Yu2019} \cite{Castro2020}. More domain-specific multi-modal datasets are being made available for tasks such as sports, traffic conditions, and videogame gameplay analysis \cite{Li2024} \cite{Xu2021} \cite{Mun2016}. The datasets for these models play a pivotal role in teaching models to detect video highlights, generate dense captions, and summarize video and auditory content.  

Videos containing tutorials and procedures also provide key content for models to learn from. Instructional videos can equip models with the skills to learn detailed temporal relationships among events and materials. Long-format tutorials specifically gauge the model's abilities to synthesize detailed explanations and lengthy procedures. They test the models’ skill at localizing key events into concise timeframes and summarizing succinctly the description of the event. Models can learn to synthesize content from demonstrations of activities and human commentary. Cooking videos have been a challenging instruction-tuning task for models. Multimodal models must learn the relationship between ingredients and measurements used, as well as preparation and recipe steps conducted to generate summaries from tutorials accurately. Models should have an understanding of different food preparation techniques, a vocabulary suitable for the cooking domain, and a temporal understanding of boundaries between different steps. 

Similar to cooking, the medical procedure domain also contains many video tutorials for users. Thousands of medical procedure tutorials are posted to YouTube by medical specialists and professionals for public use, both for educational and healing purposes. Tutorials may include best safety practices, proper equipment handling, expert recommendations, and correct techniques of various treatments, procedures, and diagnoses. These videos range from in-depth medical course material often viewed by medical students to simple tutorials and explanations for non-professionals seeking at-home diagnoses and remedies for their conditions. Both kinds of videos are excellent resources for users but often are lengthy and require considerable effort to parse through. Medical procedure videos resemble the structure of a recipe video, with the placement of ingredients and materials required for the proposed tutorial occurring before the demonstration of the main procedure at the beginning of the videos. Both domains’ demonstrations center on presenting viewers with slow play-by-play steps, narrating the exact description of each action performed in depth.

\subsection{Time-Sensitivity}
Recently, time-sensitive encoders have led to improvements in modality fusion and alignment. Before these integrations, existing Video LLMs (VLMs) could only capture global visual semantics for short clips and failed to associate the significant video content with accurate timestamps. For example, Video-LLaMA and VideoChat struggle to localize and describe meaningful events in untrimmed videos, leading to low accuracy. Additionally, the rigid compression of converting video tokens to a fixed number does not scale well to longer duration videos \cite{Maaz2023} \cite{Zhang2023a}. Neglecting video duration in these frameworks results in severe spatial-temporal semantics degradation when processing massive frames from long videos. Timestamps are versatile in utility, providing grounding and distinguishing relationships between moments. Many techniques exist within pure computer vision models for highlight detection and moment retrieval based utilizing timestamps and auditory patterns present within videos \cite{Lei2021} \cite{Zala2023} \cite{Zhang2023b} \cite{Lin2023} \cite{Yang2023}. Time-aware encoders, as used in TimeChat, enhance the binding of every frame's visual context and timestamp descriptions \cite{Ren2024}. TimeChat's sliding video Q-Former accommodates adaptive video token length during the extraction and compression of video features. The sliding Q-Former compresses frames within a sliding window allowing dynamic video token sequence generation of various lengths depending on video duration. TimeChat’s ability to unify video tokens and timestamp information allows for learning additional time-vision associations. This ability is beneficial in the comprehension of lengthy but content-packed videos. More information is encapsulated and is resistant to being left out during token sampling. This is essential when building towards generating instruction summaries, which require a significant amount of information to be parsed and synthesized without dropping content with temporal ties throughout the video. Additionally, timestamps are useful in grounding distinct temporal boundaries between steps and localizing activities to precise windows within step localization. 

\section{Approach}

\begin{figure}[ht]
    \centering
    \includegraphics[width= 8cm, height=7cm]{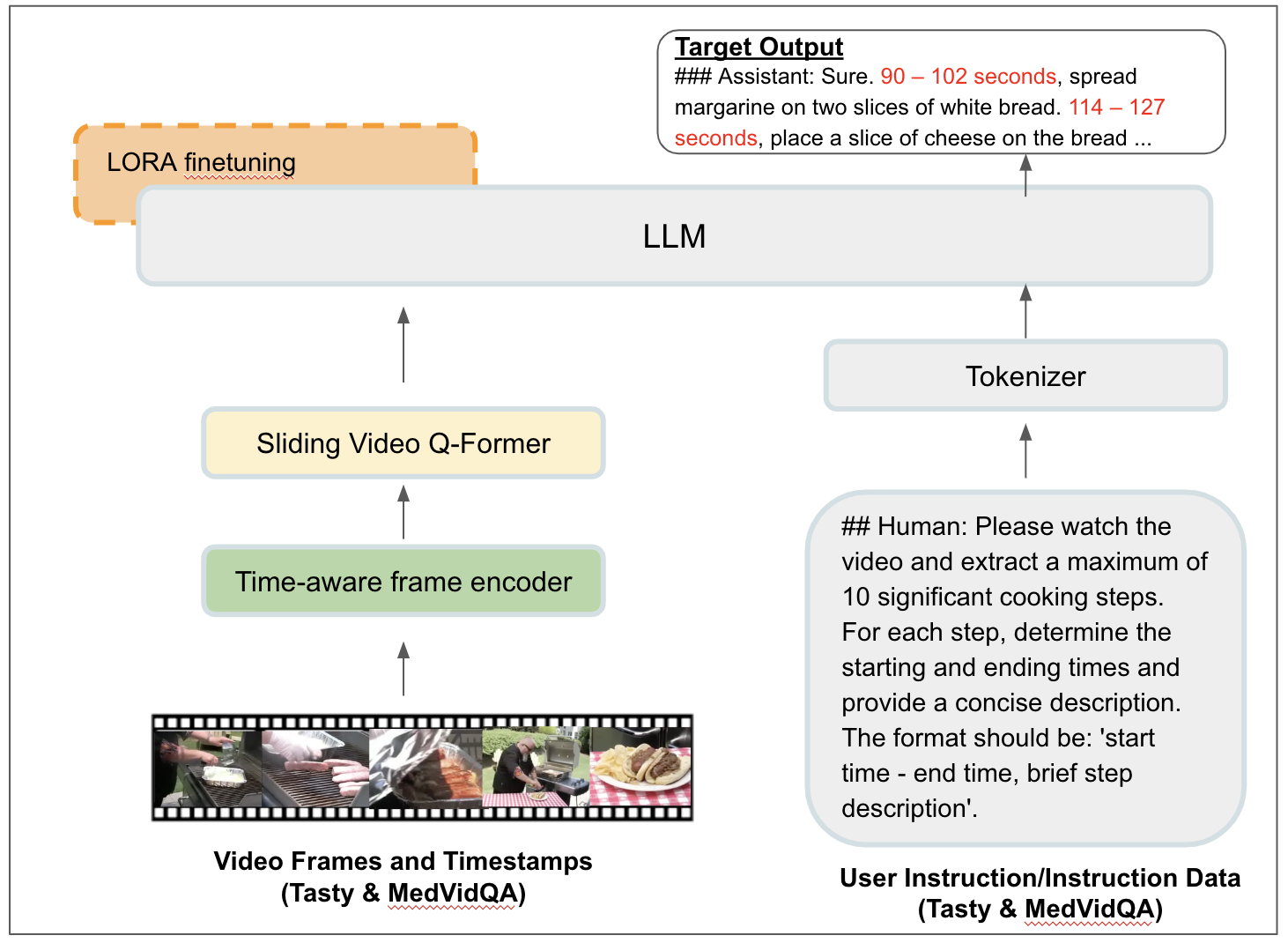}
    \caption{Finetuning approach for Domain-Expert TimeChat}
    \Description[fig:diagram]{Finetuned TimeChat extracting recipe steps}
    \label{fig:diagram2}
\end{figure}

\subsection{Data Construction}

Our approach entails finetuning TimeChat on a manually curated instruction-tuning dataset comprised of instruction data to enhance the step localization and video summarization abilities for cooking and medical procedural videos. We finetune TimeChat on restructured versions of the Tasty cooking video dataset for recipe understanding and the MedVidQA dataset published by TREC for medical procedure understanding \cite{Sener2019}. The Tasty dataset consists of 2511 recipe videos showcasing recipe procedures from an egocentric perspective. Videos do not contain instruction narration and instead display concise ingredient measurements and descriptions of steps at relevant frames.  Each recipe is annotated with the temporal boundaries for the steps within the video. This dataset is apt for training the model for better step localization, with the change of text descriptions typically serving as cues to indicate boundaries between successive steps. The MedVidQA dataset consists of 12,657 medical and health-related YouTube videos with narration of procedures. The human narration alongside the videos provides the content to synthesize into summaries. We split the Tasty and MedVidQA datasets into train and test sets used during evaluation. The Tasty train set comprises 1757 videos, while the test set comprises 754 videos. The MedVidQA train set consists of 8859 videos, while the test set contains 3798 videos. 

These two video datasets are converted into an instruction-following format to obtain high-quality video-centric instruction data. We follow the same process utilized to construct the TimeIt dataset, comprising of an instruction writing and answer formatting step \cite{Ren2024}. GPT-4 is used to extend manually written instructions into diverse expressions. We reformulate the task outputs into a user-friendly natural language response based on these written expressions. For Tasty, the original JSON files containing step annotations are restructured to include our manually written instruction queries with the labeled original timestamps and steps.

\colorlet{punct}{red!60!black}
\definecolor{background}{HTML}{EEEEEE}
\definecolor{delim}{RGB}{20,105,176}
\colorlet{numb}{magenta!60!black}

\lstdefinelanguage{json}{
    basicstyle=\normalfont\ttfamily,
    numbers=left,
    numberstyle=\scriptsize,
    stepnumber=1,
    numbersep=8pt,
    showstringspaces=false,
    breaklines=true,
    frame=lines,
    backgroundcolor=\color{background},
    literate=
     *{0}{{{\color{numb}0}}}{1}
      {1}{{{\color{numb}1}}}{1}
      {2}{{{\color{numb}2}}}{1}
      {3}{{{\color{numb}3}}}{1}
      {4}{{{\color{numb}4}}}{1}
      {5}{{{\color{numb}5}}}{1}
      {6}{{{\color{numb}6}}}{1}
      {7}{{{\color{numb}7}}}{1}
      {8}{{{\color{numb}8}}}{1}
      {9}{{{\color{numb}9}}}{1}
      {:}{{{\color{punct}{:}}}}{1}
      {,}{{{\color{punct}{,}}}}{1}
      {\{}{{{\color{delim}{\{}}}}{1}
      {\}}{{{\color{delim}{\}}}}}{1}
      {[}{{{\color{delim}{[}}}}{1}
      {]}{{{\color{delim}{]}}}}{1},
}
Below is the annotation JSON format for a video from the Tasty dataset. Notice that our reformulated task output and the original timestamp annotations refactored together to form an instructional query and labeled response. Our generated tasks generally task the model with identifying different activities and breaking down the video into segments with timestamps based on the temporal boundaries of completed activities. Descriptions should be concise and clear in describing the main focus of the corresponding segment.

\begin{lstlisting}[language=json,firstnumber=1]
{
 "video": {"Tasty/clips/PrjXoDOpJ2Y_77_355.mp4"},
 "QA": [
    {"q": "Find, identify, and determine the temporal boundaries of a series of distinct actions or steps occurring throughout the video. For each action, output the corresponding start and end timestamps, accompanied by a concise description.", 
    "a": "0.0 - 27.0 seconds, Add oil in pan.  27.0 - 67.0 seconds, Add chicken.  67.0 - 75.0 seconds, Mix sauces and garlic.  75.0 - 77.0 seconds, Add ginger and sugar.  77.0 - 80.0 seconds, Add vinegar.  80.0 - 86.0 seconds, Add soy sauce sesame oil.  86.0 - 90.0 seconds, Add water.  90.0 - 108.0 seconds, Mix it all.  108.0 - 157.0 seconds, Saute the chicken.  157.0 - 160.0 seconds, Turn heat on medium.  160.0 - 168.0 seconds, Add water chestnut.  168.0 - 186.0 seconds, Pour the sauce.  186.0 - 196.0 seconds, Cook it.  196.0 - 200.0 seconds, Turn heat off.  200.0 - 247.0 seconds, Put scallions.  247.0 - 278.0 seconds, Put in bowl."}
 ]
}
\end{lstlisting}

For the MedVidQA dataset, we construct video annotations manually through crowdsourcing, similar to methods followed by TVSum \cite{Song2015}. We utilized the Prolific platform to publish annotation requirements and collect annotator responses. Two responses were collected per video. Participants were asked to \textbf{(1)} watch the entire duration of the video \textbf{(2)} provide an importance score of 1 (not important) to 5 (very important) to each of uniform length (2 seconds) shots for the whole video. Like Tasty, these generated step annotations are restructured to include our manually written instruction queries. 

Below is an example from our constructed annotation JSON file for a video from the MedVidQA dataset. The saliency scores were collected from human annotators and normalized. 

\begin{lstlisting}[language=json,firstnumber=1]
{
 "video": {"MedVidQA/videos/takeblood.mp4"},
 "QA": [
    {"q": "Transcribed speech: None. Based on the video content and possible transcribed speech, You are given a video from the MedVidQA dataset. Please find the highlight contents in the video, determining the highlight timestamps and its saliency score on a scale from 1 to 5. The output format should be like: 'The highlight timestamps are in the 82, 84, 86, 88, 90, 92, 94, 96, 98, 100 second. Their saliency scores are 1.3, 1.7, 1.7, 1.7, 1.7, 1.3, 1.7, 2.3, 2.3, 2.3'.",
    "a": "The highlight timestamps are in the 17.0, 20.5, 37.5, 39.5, 56.5, 64.0, 73.5 seconds. Their saliency scores are 2.2, 2.2, 2.2, 3.4, 3.1, 1.9, 2.6."}
 ]
}
\end{lstlisting}

\subsection{Implementation Details}
ViT-G/14 from EVA-CLIP is used as the image encoder, and LLaMA-2 (7B) is used as the foundation language model \cite{Sun2023} \cite{Touvron2023}. The parameters for the image Q-Former are initialized from InstructBLIP’s checkpoint, while the video Q-Former is initialized from Video-LLaMA’s checkpoint. TimeChat is finetuned on our constructed instruction tuning dataset for 3 epochs using a 32 batch size with a single NVIDIA 8-V100 (32GB) GPU through LORA PEFT finetuning. All other hyperparameters and frame-level configurations follow TimeChat’s initialization.

\section{Results}

\subsection{Evaluation Setup}

We evaluate our model on the tasks of video summarization and step localization. We split the Tasty and MedVidQA datasets into train and test sets to construct some benchmarks on which to evaluate our model. The Tasty test set contains 754 videos, while the MedVidQA test set contains 3798 videos. We additionally constructed a video summarization generation dataset utilizing YouCook2 \cite{{Zhou2017}}. By default, the dataset consists of 2000 long untrimmed videos from 89 cooking recipes; each distinct recipe has 22 videos on average. The procedure steps for each video are annotated with temporal boundaries and are described by imperative sentences. The videos were downloaded from YouTube and are all in the third-person viewpoint. All the videos are unconstrained and can be performed by individuals at their houses with unfixed cameras. We parsed through the associated recipes from each video to create a dataset of questions asking for summaries for each recipe procedure. Our final dataset contained 408 videos from 62 different cooking recipes. The MedVidQA and YouCook2 datasets evaluate video summarization abilities across tested models, while Tasty is used in evaluating step localization. Hence, for MedVidQA and YouCook2, we report the maP and f1-scores per video. For Tasty, we report the maP score. We compare our model to TimeChat before finetuning, Video-LLaMA, and Valley. All compared models are considered end-to-end models, directly taking in videos as inputs and generating text outputs in an end-to-end matter. 

\subsection{Results Analysis}

To expand further on our findings from finetuning TimeChat for domain-specific video summarization and step localization enhancement, we delve deeper into the specific outcomes observed through our experiments. Our investigation centered on comparing the quality of generated procedural video summarizations and step understanding of baseline TimeChat and TimeChat finetuned on the video domains of recipes and medical procedures. Our experimental results demonstrate that our method of finetuning TimeChat leads to some boosts in generating precise video summarizations and distinguishing successive steps from one another over general-purpose trained TimeChat. This improvement is quantitatively evident when examining the f1 score comparisons of finetuned TimeChat (FT-TimeChat) and baseline TimeChat on the constructed YouCook2 datasets in Table 1. The model subjected to finetuning performed with a slightly higher f-1 score than baseline TimeChat as well as Video-LLaMA and Valley. See Table 1 for Video Summarization comparisons across evaluated models. 

\begin{table}[ht]
    \centering
    \begin{tabular}{lccccc}
        \toprule
        \multirow{1}{*}{Model} & \multicolumn{2}{c}{YouCook2} & \multicolumn{2}{c}{MedVidQA} \\
        \cmidrule(r){2-3} \cmidrule(r){4-5}
        & f1 & maP & f1 & maP \\
        \midrule
        Video-LLaMA & 2.6 & 11.3 & 1.4 & 8.7 \\
        Valley & 1.9 & 10.9 &  1.2 & 7.4 \\
        TimeChat & 12.4 & 17.6 & 6.9 & 14.5 \\
        FT-TimeChat & 12.7 & 16.8 & 7.3 & 13.7 \\
        \bottomrule
    \end{tabular}
    \caption{Performance of end-to-end models on video summarization of videos from YouCook2 and MedVidQA.}
    \label{tab:performance}
\end{table}

Table 2 compares step localization across each model on the Tasty test dataset. Mean average precision (maP) was the only metric tracked in evaluating step localization. 

\begin{table}[ht]
    \centering
    \begin{tabular}{lccccc}
        \toprule
        \multirow{1}{*}{Model} & \multicolumn{2}{c}{Tasty}  \\
        \cmidrule(r){2-3} & maP \\
        \midrule
        Video-LLaMA & 15.6 \\
        Valley & 15.2 \\
        TimeChat & 17.7 \\
        FT-TimeChat & 18.1 \\
        \bottomrule
    \end{tabular}
    \caption{Performance of end-to-end models on step localization of videos from Tasty.}
    \label{tab:performance2}
\end{table}

Despite finetuning on a larger corpus of medical-related videos than cooking videos given our train and test set sizes mentioned earlier, f1 scores and mean average precision scores on YouCook2 and MedVidQA seem to indicate that the model performed better on video summarization within the cooking procedure domain. When evaluating step localization abilities on the Tasty dataset, fine-tuned TimeChat slightly outperforms the baseline version in separating out steps within recipes. 

\begin{figure}[ht]
    \centering
    \includegraphics[width=8.5cm, height=7cm]{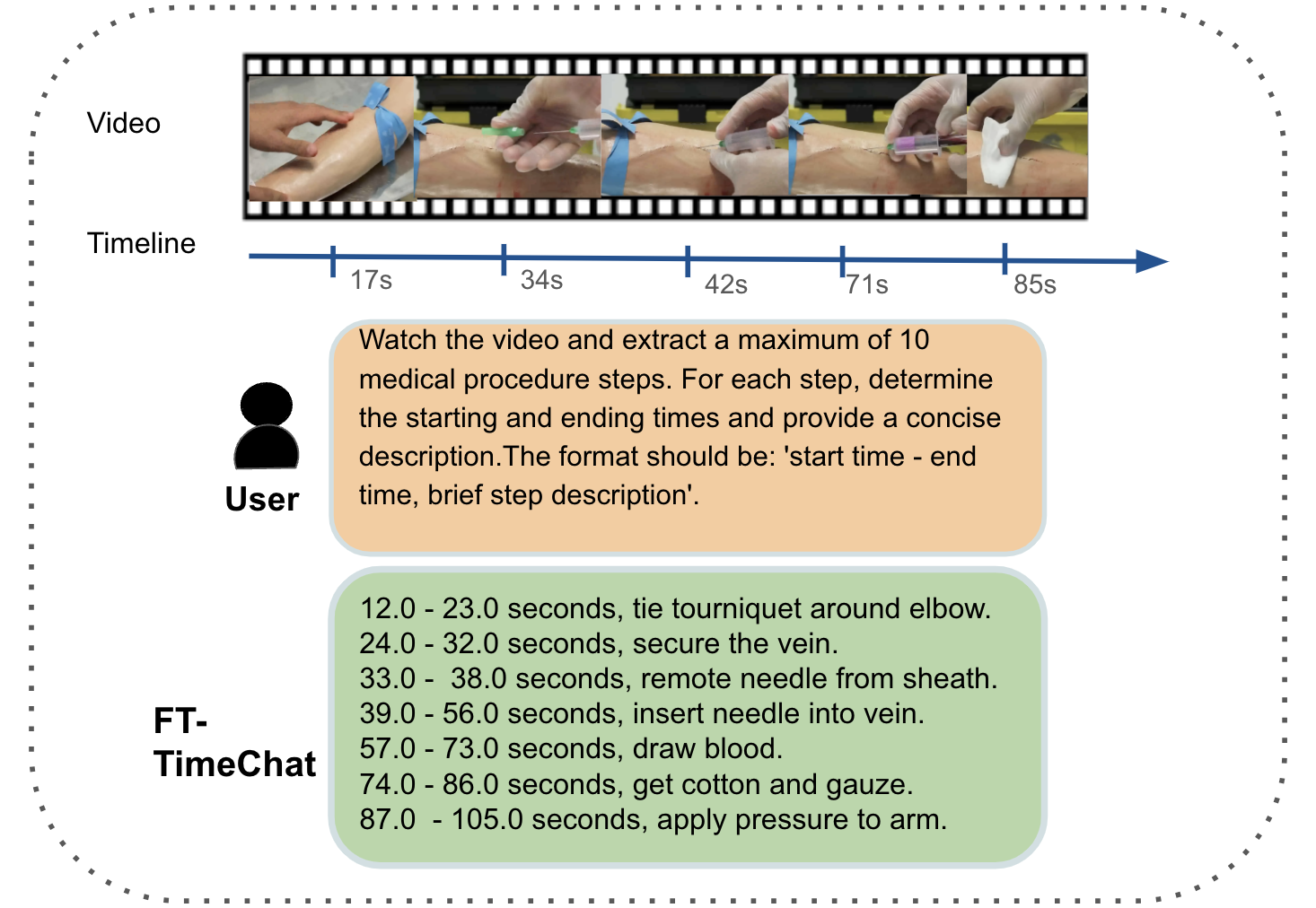}
    \caption{Finetuned TimeChat on Medical Domain Video Summarization}
    \Description[fig:diagram]{Finetuned TimeChat extracting recipe steps}
    \label{fig:diagram3}
\end{figure}

One observation we made that could explain this finding was that baseline TimeChat tended to mistake body parts showcased in the video during a demonstration. Before finetuning, TimeChat frequently mistook an arm for a leg in a video demonstrating how to draw blood from patients. Certain medical equipment and materials like tourniquets and gauze were often mislabeled in summaries due to such terminology not being included in the model's vocabulary. Finetuning expanded TimeChat's vocabulary subject to the domain, helping produce more precise and descriptive summaries. Finetuned TimeChat also tended to produce more concise recipe steps than baseline TimeChat, whose responses were often more lengthy.

There is room for more robust evaluation through employing human evaluation on a subset of responses from finetuned TimeChat and baseline TimeChat. We can gather a group of people to blindly judge between the summaries generated by both models which response is more accurate to a provided ground truth for a subset of responses produced by both models. Additionally, we can also break down the evaluation into 2 stages: \textbf{(1)} where annotators can check if generated segments line up change of events or change of activities within the video and \textbf{(2)} annotators can check if the generated description aligns with the activity occurring during the segment.

\section{Conclusion and Future Work}

\subsection{Conclusion}
In conclusion, this research underscores the potential of leveraging Multimodal Large Language Models to enhance the comprehension and summarization of instructional videos with the cooking and medical doamins. By fine-tuning TimeChat on domain-specific datasets like Tasty and MedVidQA, we have demonstrated the model's improved ability to localize key events and generate concise, accurate summaries of step-by-step procedures in cooking and medical tutorials. The restructuring of domain-specific datasets into an instruction-following format, combined with the use of GPT-4 for expanding instruction diversity, has facilitated the creation of high-quality video-centric instruction data. This data is crucial for training models to handle a variety of instructional queries accurately and effectively. The enhanced capabilities of TimeChat in understanding and summarizing video content hold significant promise for practical applications. From providing detailed cooking instructions to summarizing complex medical procedures, the model can offer personalized, step-by-step guidance tailored to the unique aspects of each scenario.

\subsection{Future Work}
One avenue of further investigation is training the frame encoder to segment videos into non-continuous time frames. Currently, generated segments occur continuously with the time frames of successive steps often having start times that line up immediately after the end time of prior steps. While this is useful for sectioning of a video into chunks based on steps or activities, it is also useful to generate tight segments that are confined to the key highlights. Currently, segments can be edited down to tighter time frames that encapsulate content aligned with the exact descriptions generated by the model. If each generated segment were to be combined into a new video, the video should only contain the video highlights. 

Another area worth exploring is the use of medical procedure data from other modalities to finetune the model on more domain-specific information. There was a significant amount of data from various datasets available within the cooking domain that could be used in this experiment for finetuning the model. Within the medical procedure domain, we were limited to MedVidQA to provide videos that could be used in both training and evaluation. One possibility to add more medical data could be generating instruction data from health articles that provide tutorials and at-home procedures. Many at-home remedies and healthy-related articles with images demonstrating procedures exist online on websites like MayoClinic, MedlinePlus, and WikiHow. Through careful processing of required materials, procedural steps, and recommendations, we can construct additional datasets that can expose the model to a larger corpus of medical terminology and practices useful in growing the model's reasoning within the medical domain.

From a development perspective, we hope to containerize our current workflow that runs on a 32GB 8-V100 GPU cluster hosted on an AWS Lambda instance. This server instance has been quite useful throughout this project but has also incurred quite the cost. We wish to create a Docker image of our development environment to simplify the infrastructure setup process for different server architectures. We also hope to look into more fine-grained batch processing at the fine-tuning stage to speed up the process. Additionally, we hope to build a user interface with Gradio to accommodate the display of the generated segment summarizations, along with their corresponding video frame groundings. Presenting users with an edited down version of the video from the segments along with the segment descriptions as captions to accompany the video would be useful in application of this type of model.

\balance

\end{document}